\documentclass[pdflatex,sn-mathphys-num]{sn-jnl}
\usepackage{changes}
\usepackage{graphicx}%
\usepackage{multirow}%
\usepackage{amsmath,amssymb,amsfonts}%
\usepackage{amsthm}%
\usepackage{mathrsfs}%
\usepackage{booktabs}
\usepackage{xcolor}%
\usepackage{soul}
\usepackage{textcomp}%
\usepackage{manyfoot}%
\usepackage{booktabs}%
\usepackage{algorithm}%
\usepackage{algorithmicx}%
\usepackage{algpseudocode}%
\usepackage{listings}%
\usepackage{times}
\usepackage{latexsym}
\usepackage{multirow}
\usepackage{microtype}
\usepackage{hyperref}
\usepackage{url}
\usepackage{booktabs}
\usepackage{multirow}
\usepackage{algorithm}
\usepackage{algpseudocode}
\usepackage{lineno}
\usepackage{physics}
\usepackage{environ}
\usepackage{subcaption}

\usepackage{float}

\usepackage{tcolorbox}
\usepackage{microtype}
\usepackage{enumitem}
\usepackage{subfiles}
\usepackage{hyperref}
\usepackage{inconsolata}
\usepackage{array} 
\usepackage{booktabs}

\theoremstyle{thmstyleone}%
\theoremstyle{thmstyletwo}%

\theoremstyle{thmstylethree}%

\begin{document}

\title[Article Title]{WiseMind: A Knowledge-Guided Multi-Agent Framework for Accurate and Empathetic Psychiatric Diagnosis}

\author[1,2]{\fnm{Yuqi} \sur{Wu}}\email{ywu\_bh@fudan.edu.cn}
\equalcont{These authors contributed equally to this work.}

\author[3]{\fnm{Guangya} \sur{Wan}}\email{wxr9et@virginia.edu}
\equalcont{These authors contributed equally to this work.}

\author*[4]{\fnm{Jingjing} \sur{Li}}\email{jingjing.li@virginia.edu}

\author[1,2]{\fnm{Shengming} \sur{Zhao}}

\author[1]{\fnm{Lingfeng} \sur{Ma}}

\author[1]{\fnm{Tianyi} \sur{Ye}}

\author[5]{\fnm{Mike} \sur{Zhang}}

\author[5]{\fnm{Ion} \sur{Pop}}

\author*[5]{\fnm{Yanbo} \sur{Zhang}}
\email{yanbo9@ualberta.ca}

\author*[1,2]{\fnm{Jie} \sur{Chen}}
\email{jc65@ualberta.ca}

\affil*[1]{\orgdiv{College of Biomedical Engineering}, 
\orgname{Fudan University}, 
\orgaddress{\city{Shanghai}, \postcode{200433}, \country{China}}}

\affil[2]{\orgdiv{Department of Electrical and Computer Engineering}, 
\orgname{University of Alberta}, 
\orgaddress{\city{Edmonton}, \state{Alberta}, \postcode{T6G 2V4}, \country{Canada}}}

\affil[3]{\orgdiv{School of Data Science}, 
\orgname{University of Virginia}, 
\orgaddress{\city{Charlottesville}, \state{Virginia}, \postcode{22903}, \country{USA}}}

\affil[4]{\orgdiv{McIntire School of Commerce}, 
\orgname{University of Virginia}, 
\orgaddress{\city{Charlottesville}, \state{Virginia}, \postcode{22903}, \country{USA}}}

\affil[5]{\orgdiv{Department of Psychiatry}, 
\orgname{University of Alberta}, 
\orgaddress{\city{Edmonton}, \state{Alberta}, \postcode{T6G 2R3}, \country{Canada}}}


\abstract{
Large Language Models (LLMs) offer promising opportunities to support mental healthcare workflows, yet they often lack the structured clinical reasoning needed for reliable diagnosis and may struggle to provide the emotionally attuned communication essential for patient trust. Here, we introduce WiseMind, a novel multi-agent framework inspired by the theory of Dialectical Behavior Therapy designed to facilitate psychiatric assessment. By integrating a "Reasonable Mind" Agent for evidence-based logic and an "Emotional Mind" Agent for empathetic communication, WiseMind effectively bridges the gap between instrumental accuracy and humanistic care. Our framework utilizes a Diagnostic and Statistical Manual of Mental Disorders, Fifth Edition (DSM-5)-guided Structured Knowledge Graph to steer diagnostic inquiries, significantly reducing hallucinations compared to standard prompting methods. Using a combination of virtual standard patients, simulated interactions, and real human interaction datasets, we evaluate WiseMind across three common psychiatric conditions. WiseMind outperforms state-of-the-art LLM methods in both identifying critical diagnostic nodes and establishing accurate differential diagnoses. Across 1206 simulated conversations and 180 real user sessions, the system achieves 85.6\% top-1 diagnostic accuracy, approaching reported diagnostic performance ranges of board-certified psychiatrists and surpassing knowledge-enhanced single-agent baselines by 15–54 percentage points. Expert review by psychiatrists further validates that WiseMind generates responses that are not only clinically sound but also psychologically supportive, demonstrating the feasibility of empathetic, reliable AI agents to conduct psychiatric assessments under appropriate human oversight.}


\maketitle

\section*{Introduction}\label{sec1}

Psychiatric assessment and diagnosis are among the most demanding tasks in clinical medicine, requiring clinicians to synthesize verbal symptom histories, behavioral observations, and contextual factors within an empathic, adaptive interview \citep{Carlat2023,nordgaard2013psychiatric}. Clinicians must differentiate overlapping symptom clusters, apply temporal qualifiers and exclusion rules, monitor for self-harm or crisis risk, and simultaneously maintain rapport and respond to emotional cues \citep{First2024DSM5TR,Demazeux2015DSM5,Carlat2023,nordgaard2013psychiatric}. These challenges are compounded by escalating clinical demand \citep{Kessler2005PrevalenceNCS}, workforce shortages \citep{thomas2006continuing,butryn2017shortage}, long training pipelines \citep{das2023graduate,button2025clinical}, and expectations for culturally attuned, equitable care \citep{topol2019high,moon2023ethical}. As these pressures intensify, there is growing interest in whether large language models (LLMs) could assist intake, triage, and early-stage diagnostic reasoning.

Although LLMs have demonstrated promising performance on medical board examinations \citep{Smith2023LLMReview,Johnson2023HealthcareLLM}, most clinical Natural Language Processing (NLP) systems are not yet fully aligned with the cognitive, interpersonal, and ethical demands of psychiatric assessment. Many rely on pretraining, fine-tuning, in-context learning (ICL), or retrieval-augmented generation (RAG) \citep{wu2023systematic,xue2024ai,demetriou2020machine,wu2023automatic,freidel2025knowledge}, yet they are evaluated mainly with technical metrics such as F1-score or Bilingual Evaluation Understudy (BLEU). As highlighted in recent clinical NLP surveys \citep{Wu2022UKClinicalNLP,Croxford2025LLMEval}, these systems often perform well in silico but are still insufficient to meet real-world expectations for systematic diagnostic reasoning, flexible interviewing, empathic engagement, and stringent safety supervision \citep{tu2025towards}. Three gaps restrict their clinical utility, as summarized in Table \ref{tab:design-gap}.

\begin{table}[ht]
\centering
\caption{Interdisciplinary contextualization framework showing gaps in current approaches and WiseMind’s corresponding solutions for psychiatric differential diagnosis.}
\label{tab:design-gap}

\begin{tabular}{>{\raggedright\arraybackslash}p{2cm} >{\raggedright\arraybackslash}p{2cm} >{\raggedright\arraybackslash}p{3cm} >{\raggedright\arraybackslash}p{4cm}}
\hline
\textbf{Design Gaps} & \textbf{Existing Methods} & \textbf{Observed Limitations} & \textbf{WiseMind Design Component} \\
\hline
\vspace{0pt}
\textbf{Knowledge Integration} & 
\vspace{0pt}Pretraining \citep{Wang2023PretrainedModels}, ICL \citep{dong-etal-2024-survey}, RAG \citep{yoran2024making}& 
\begin{itemize}[leftmargin=*, nosep]
    \item Flat, unstructured knowledge
    \item Lack of clinical decision paths
    \item Reactive behavior
\end{itemize} & 
\vspace{0pt}Structured Knowledge--Guided Proactive Reasoning
\begin{itemize}[leftmargin=*, nosep]
    \item DSM-guided state graph \citep{Demazeux2015DSM5}
    \item Action-based transitions
    \item Structured, proactive dialog
\end{itemize} \\
\hline
\vspace{0pt}\textbf{Process Modeling} & 
\vspace{0pt}Single-agent LLMs \citep{singhal2023large,chung2024scaling}; heuristic multi-agent setups \citep{kim2024mdagents,tu2025towards} & 
\begin{itemize}[leftmargin=*, nosep]
    \item Overemphasis on reasoning
    \item Shallow emotional responses
    \item Lack of theoretical grounding
\end{itemize} & 
\vspace{0pt}Theory-Informed Dual-Agent Architecture
\begin{itemize}[leftmargin=*, nosep]
    \item Dual-agent design grounded in DBT \citep{linehan1993cognitive}
    \item RA for diagnosis; EA for empathy
\end{itemize} \\
\hline
\vspace{0pt}\textbf{Evaluation Scope} & 
\vspace{0pt}Accuracy metrics (e.g., F1, BLEU) \citep{wu2023systematic,xue2024ai,demetriou2020machine} & 
\begin{itemize}[leftmargin=*, nosep]
    \item Ignores safety, empathy, trust
    \item No user-centred or ethical testing \citep{aggarwal2024cultural,kanjee2023accuracy}
\end{itemize} & 
\vspace{0pt}Multi-Faceted Evaluation Strategy
\begin{itemize}[leftmargin=*, nosep]
    \item Multi-tier evaluation: user, expert, ethical
    \item Adversarial and human-in-the-loop tests \citep{topol2019high,moon2023ethical,kerz2023toward}
\end{itemize} \\
\hline
\end{tabular}
\end{table}

First, a domain knowledge gap limits diagnostic reliability. Psychiatric diagnosis is governed by highly structured decision pathways defined in the Diagnostic and Statistical Manual of Mental Disorders, Fifth Edition (DSM-5) \citep{american2013dsm5,regier2013dsm,First2024DSM5TR,Demazeux2015DSM5} and the International Classification of Diseases, Eleventh Revision (ICD-11) \citep{who2019icd}. These frameworks organize disorders into hierarchical symptom clusters, supplemented by temporal qualifiers, exclusion criteria, and specifiers that guide clinicians in distinguishing among more than 150 frequently overlapping conditions. This structured clinical reasoning process—known as psychiatric differential diagnosis (DDx)—is essential for determining the most accurate and safe diagnostic formulation \citep{Demazeux2015DSM5}. Because DDx relies almost entirely on verbal histories rather than laboratory or imaging findings, effective diagnostic-support systems must adhere to these structured pathways to maintain clinical coherence \citep{Carlat2023,nordgaard2013psychiatric}. Yet current LLM knowledge-integration methods represent information as flat or lightly tagged text \citep{yoran2024making}, overlooking this tree-like structure and producing reactive, prompt-driven behavior that increases the risk of omissions, disorganized dialogue, and missed exclusion criteria \citep{lu2024multimodal,meurisch2020exploring,10.1145/3560815}.

Second, a process gap prevents empathic and adaptive interviewing. Effective psychiatric assessment requires balancing analytic reasoning and empathic engagement \citep{nordgaard2013psychiatric}. Clinicians build rapport, attune to affect, sense hesitation, respond to emotional cues, and dynamically adjust questioning. These relational skills strongly shape disclosure, including suicidality, trauma, mania, psychosis, or substance use \citep{savander2024take}. This balance mirrors Dialectical Behavior Therapy (DBT)’s distinction between the “reasonable mind” (cognitive, rule-based) and the “emotional mind” (affective, validating) \citep{linehan1993cognitive}. Current LLMs excel at chain-of-thought reasoning \citep{singhal2025toward,chung2024scaling,singhal2023large} but can sometimes lack thoughtful emotional responses. By contrast, empathy-oriented systems such as Woebot, Wysa, and EmoGPT foster rapport but may not fully reflect the clinical reasoning \citep{fitzpatrick2017delivering,inkster2018empathy,lan2024depressiondiagnosisdialoguesimulation}. Emerging multi-agent systems \citep{kim2024mdagents,tu2025towards,mcduff2025towards} begin to integrate these capabilities but still rely on heuristic or opaque coordination mechanisms rather than psychologically grounded theories and practices.

Third, an evaluation gap limits safe deployment. High-stakes psychiatric AI requires robust assessment of empathy, conversational quality, trustworthiness, fairness, and crisis-response behavior—not only numerical accuracy \citep{aggarwal2024cultural,kanjee2023accuracy,Yu2022EmpathyDevelopment,Licciardone2024EmpathyChronicPain}. Yet fewer than 15\% of mental health NLP studies incorporate user-centered or ethical metrics \citep{aggarwal2024cultural,kanjee2023accuracy}. Few systematically probe resilience to adversarial prompts or suicidal ideation \citep{robertson2023diverse,kerz2023toward,pashak2022build}. Ethical evaluations are particularly challenging because they require interacting with vulnerable populations, raising risks of harm and participatory injustice \citep{ferrara2022MLpsychosis,Pozzi2025ParticipatoryInjustice,Meadi2025ConversationalAI} and requiring intensive human oversight \citep{BearDontWalk2022ClinicalNLPEthics,Zhang2022MentalIllnessNLP}.  Existing approaches therefore lack the breadth and depth of evaluation required for clinical adoption \citep{tu2024multiple}.

Addressing these gaps, we introduce WiseMind, a multi-agent LLM framework inspired by the “reasonable mind” (rational, cognitive) and “emotional mind” (affective, intuitive) constructs of DBT \citep{linehan1993cognitive}. WiseMind operationalizes a “contextualization trio” across \textit{knowledge}, \textit{process}, and \textit{evaluation} layers and integrates the complementary strengths of analytic reasoning and empathic communication through a theory-informed dual-agent architecture. WiseMind is explicitly designed as an \textit{assistive} technology for intake and triage support, rather than a stand-alone diagnostic system. It operationalizes this “contextualization trio” through three coordinated components:
(i) Structured Knowledge--Guided Proactive Reasoning, which encodes the full DSM-5 decision graph as a state-transition knowledge graph to steer criterion-aligned question sequencing; 
(ii) Theory-Informed Dual-Agent Architecture, which deploys a DBT-aligned workflow wherein a \emph{Reasonable-Mind Agent (RA)} consults the graph to determine the next diagnostic action while an \emph{Emotional-Mind Agent (EA)} re-expresses that intent in empathic, trust-building language; and 
(iii) Multi-Faceted Evaluation Strategy, which couples technical testing with a three-tier validation pipeline—simulated patients, lay-user studies, and expert clinician review—further strengthened by ethical stress-testing for self-harm scenarios and bias audits across age and gender subgroups.

We evaluate WiseMind on three common mental health conditions—depressive mood (depression), elevated mood (hypomania or mania), and anxious mood (anxiety). Across 1206 simulated conversations and 180 real user sessions, the system achieves 85.6\% top-1 diagnostic accuracy, surpassing knowledge-enhanced single-agent baselines by 15–54 percentage points (p $<$ 0.01). In parallel, blinded raters judged WiseMind’s empathic quality 13\%--54\% higher and its advice precision 12\%--30\% higher than competing models (ICC2 $>$ 0.75). Ethical audits further show that the system refuses unsafe instructions, flags suicidal ideation, and reduces hallucinated medication recommendations.

This study makes several contributions to digital psychiatry and the design of clinically reliable LLM systems. First, we encode the DSM-5 differential-diagnosis pathways into a structured state-transition knowledge graph, enabling proactive, criterion-aligned interviewing rather than reactive text retrieval. Second, we introduce a DBT-informed dual-agent architecture that cleanly separates analytic diagnostic reasoning from empathic, patient-centered communication—a requirement often unmet by single-agent LLMs. Third, we develop a comprehensive human-centered evaluation pipeline that spans virtual patient interactions, real-user feedback, clinician review, and ethical stress-testing for suicidality and demographic bias. Finally, we demonstrate that WiseMind delivers clinically meaningful improvements in diagnostic accuracy,  empathetic performance, and medical proficiency, illustrating how deep domain contextualization across knowledge, process, and evaluation can inform future work on assistive LLM systems in psychiatric and other high-stakes domains.

\section*{Results}\label{sec2}
\subsection*{System Overview}
\begin{figure*}[ht!]
    \centering
    \includegraphics[width=\linewidth]{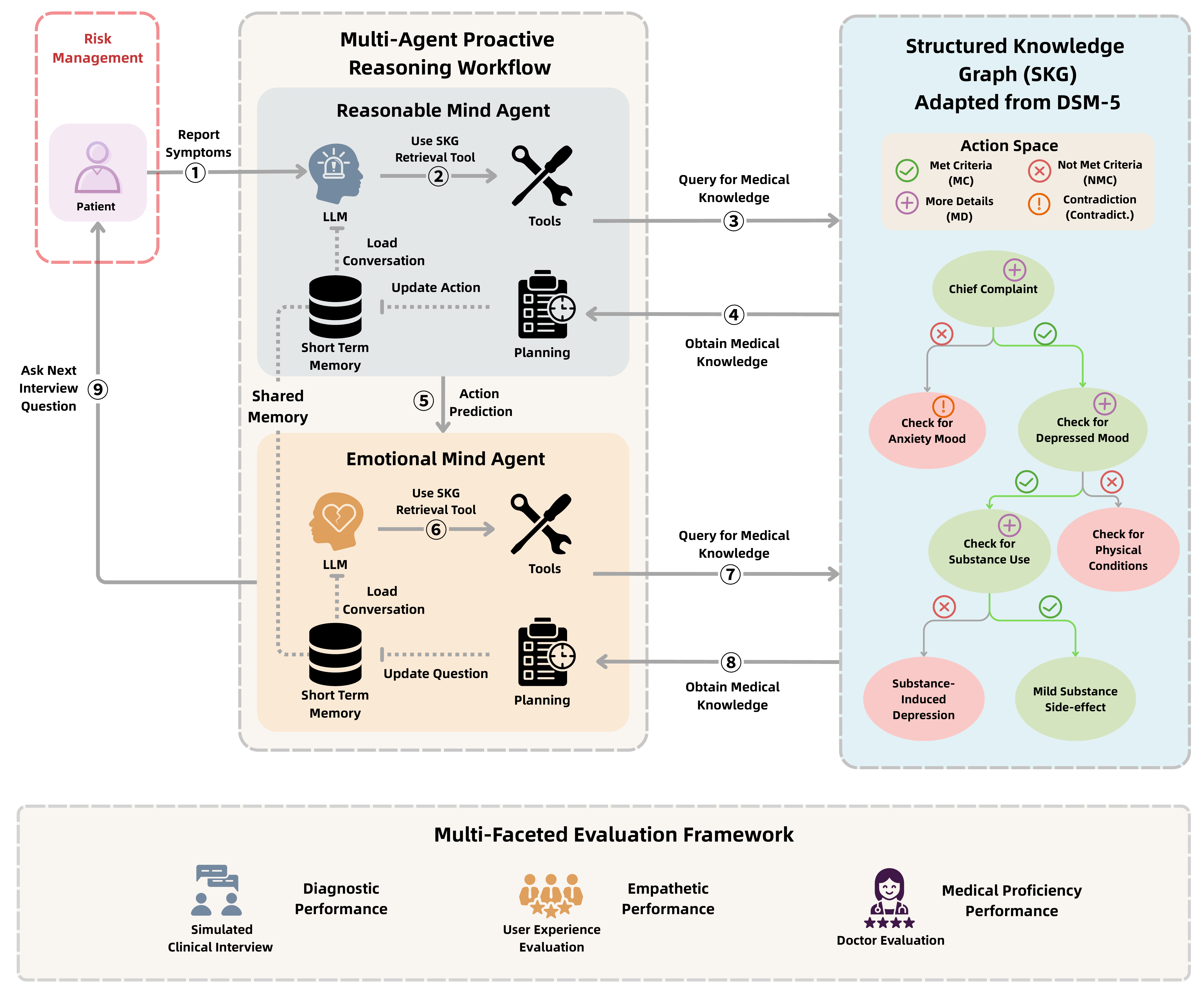}
    \caption{Overview of the WiseMind system. A DSM-5–derived structured knowledge graph (SKG) guides the Reasonable Mind Agent (RA) in performing structured knowledge retrieval (tool usage), LLM-based reasoning (planning), and action prediction. The Emotional Mind Agent (EA) translates (planning) this diagnostic intent into empathic, patient-facing responses based on retrieved medical information (tool usage). Both agents share a short-term memory that tracks conversation state, symptom evolution, and risk signals. A risk-management layer filters unsafe or contradictory content. System performance is evaluated through a multi-tier framework combining simulated interviews, human-participant interactions, and psychiatrist ratings.}

    \label{fig:workflow}
\end{figure*}

To address the design gaps identified in the Introduction, we propose WiseMind. Fig.~\ref{fig:workflow} illustrates the overall WiseMind Workflow, demonstrating how the system transforms patient input into clinically validated diagnostic inquiries.

WiseMind employs two complementary agents—the Reasonable Mind Agent (RA) and the Emotional Mind Agent (EA)—which synchronize through a Shared Short-Term Memory module to track conversation state and symptom evolution. The workflow proceeds step-by-step, corresponding to the numbered flow in the figure. The process begins when the patient reports symptoms (1). This input is screened by the \textit{Risk Management} layer to detect high-risk language (e.g., self-harm) before entering the diagnostic loop. Then from steps 2–5, the RA serves as the primary interface. It processes the input and uses the SKG Retrieval Tool (2) to query (3) and obtain (4) relevant medical context from the SKG. Based on this, the RA generates an initial \textit{Action Prediction} (5) (i.e., whether to ask for more details or validate criteria) within the defined Action Space. Steps 6–9 was consist of Emotional Processing and Question Formulation. The action intent from the RA is passed to the EA to generate patient-facing communication (step 5). The EA validates the intent against DSM-5 clinical logic by using SKG tool (6) to query (7) and obtain (8) medial knowledge. It then applies therapeutic language guidelines—such as validation, reflective phrasing, and non-judgmental tone—to maintain empathy. Finally, it integrates these clinical and emotional cues to formulate the next interview question, completing the Ask Next Interview Question (9) step.

Both agents operate over the DSM-5-adapted SKG, ensuring that every conversational turn is grounded in established psychiatric criteria while maintaining interactional fluidity. For detailed examples of conversations and a step-by-step breakdown of how different parts of the agentic workflow contribute to the clinical dialogue, see Supplementary Section 1.6.

\begin{figure}[ht!]
    \centering
    \includegraphics[width=\linewidth]{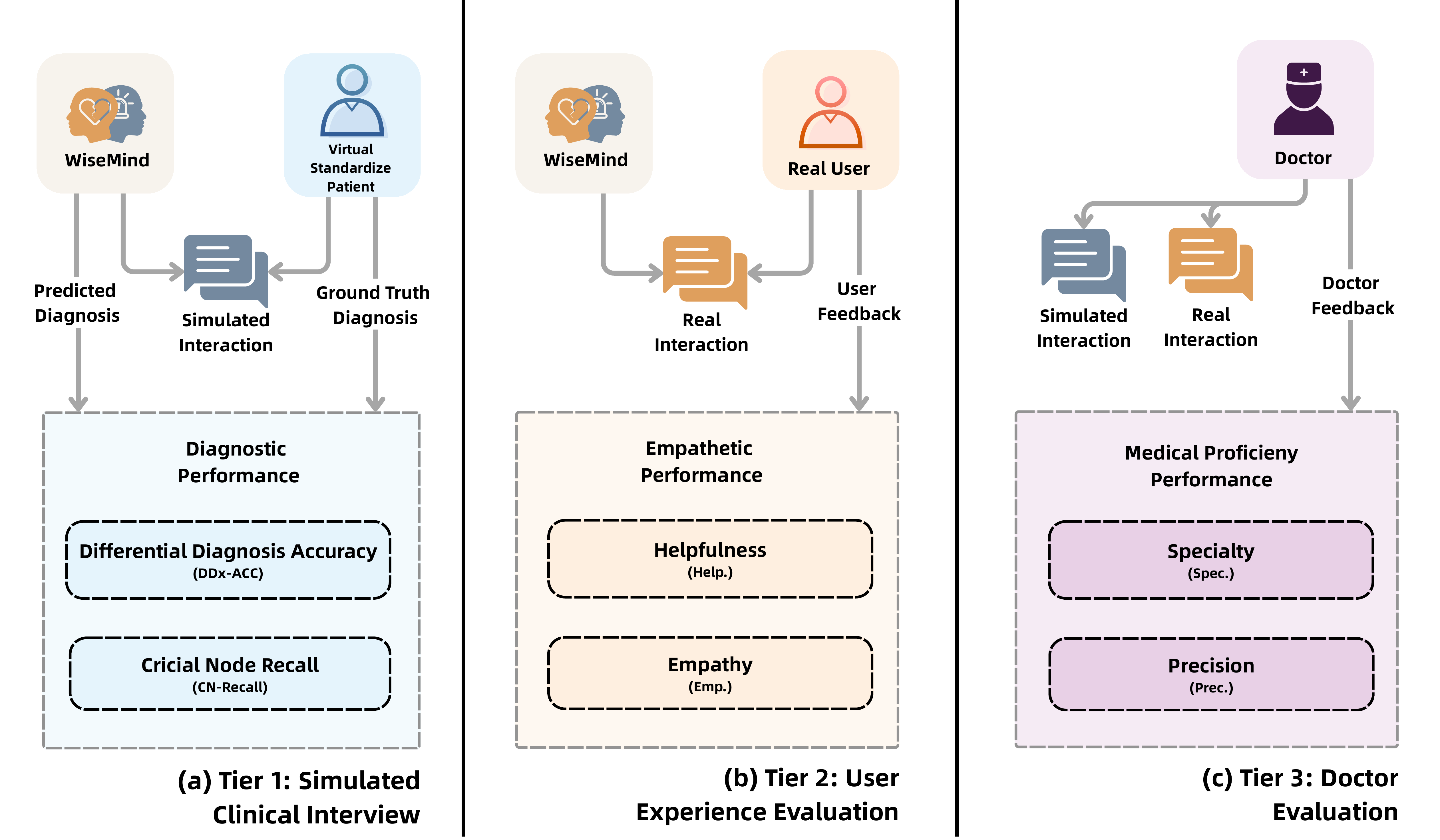}
    \caption{Multi-Faceted Evaluation Framework for WiseMind. The system's performance is holistically validated across three distinct tiers, ensuring the system meets standards for technical accuracy, patient experience, and medical proficiency:(a) \textbf{Tier~1 (Simulated Interaction Evaluation)}: Simulated Clinical Interview assesses Diagnostic Performance by comparing WiseMind's Predicted Diagnosis against the Ground Truth Diagnosis of a Virtual Standardized Patient. Metrics quantified include Differential Diagnosis Accuracy (DDx-ACC) and Critical Node Recall (CN-Recall). (b) T\textbf{ier~2 (Real Interaction Evaluation)}: User Experience Evaluation assesses Empathetic Performance via real-time interaction with a Real User. Metrics are based on User Feedback and quantify the system's Helpfulness (Help.) and Empathy (Emp.).(c) \textbf{Tier~3 (Cross-Setting Evaluation)}: Doctor Evaluation engages medical professionals who review conversation transcripts from both simulated and real interactions to assess Medical Proficiency Performance based on clinical standards. Metrics include Specialty (Spec.) and Precision (Prec.).}

    \label{fig:evaluation-framework}
\end{figure}

\subsection*{Dataset and Experiment Setup}
To assess whether WiseMind’s design components translate into measurable improvements in 
psychiatric diagnostic performance, we conduct a comprehensive, multi-faceted evaluation 
aligned with the interdisciplinary gaps identified in the Introduction. Conventional 
LLM-based diagnostic evaluations typically emphasize benchmark accuracy or single-turn 
reasoning, but psychiatric interviewing requires a broader assessment framework that 
captures diagnostic rigor, conversational rapport, clinical credibility, and safety. 
Guided by these requirements, we evaluate WiseMind across three complementary tiers 
(Fig.~\ref{fig:evaluation-framework}), each reflecting a core competency of real-world 
psychiatric practice.

\begin{table}[h!]
\centering
\caption{{Summary of datasets used in the WiseMind evaluation pipeline.}}
\label{tab:dataset-overview}
\begin{tabular}{p{1.5cm} p{4cm} p{4cm} p{2cm}}
\hline
\textbf{Dataset} & \textbf{Source and Description} & \textbf{Purpose in Evaluation} & \textbf{Size} \\
\hline
\textbf{VSP Corpus (generated)} & DSM-driven and human-grounded synthetic patient profiles & Structured, reproducible diagnostic test scenarios & 192 profiles \\
\hline
\textbf{Simulated Interaction Corpus} & Multi-turn dialogues between WiseMind and all VSPs & Quantitative evaluation of diagnostic reasoning & 1,206 dialogues \\
\hline
\textbf{Real Human Interaction Corpus} & Human actors retelling VSP stories with natural variation & Ecological validity; user and clinician evaluations & 180 sessions \\
\hline
\end{tabular}
\end{table}

The Tier~1 evaluation, namely the Simulated Clinical Interview, was designed to evaluate the diagnostic performance of the WiseMind system. We evaluate diagnostic reliability using \emph{Virtual Standardized Patients (VSPs)} \citep{maicher2017developing,hubal2000virtual,reger2021virtual,wu2023automatic,wu2024callm}, which are DSM-derived, multi-turn patient profiles designed for structured yet naturalistic interactions. Each VSP is paired with a DSM-5–aligned ground-truth diagnosis and symptom narrative (Supplementary Table 1). The evaluation spans 24 depressive, 15 bipolar, and 25 anxiety leaf diagnoses encoded in the SKG. For each diagnostic leaf, we generate three VSP variants, yielding 192 simulated patient profiles reflecting realistic variability in symptom presentation (VSP Corpus).

During evaluation, VSPs interact with WiseMind via three standard symptom-entry pathways—\emph{depressed mood}, \emph{anxiety}, and \emph{substance use}—following the proactive flow described in Fig.~\ref{fig:evaluation-framework}, yielding 1,206 simulated, multi-turn doctor–patient interactions (Simulated Interaction Corpus). These dialogues contain the complete conversation transcript, the RA's action sequence, the EA's phrasing behavior, and all intermediate updates to the SKG. This dataset enables quantitative assessment of diagnostic accuracy, critical node recall, hallucination frequency, reasoning-path adherence, and characteristic failure modes. For each dialogue, WiseMind outputs a predicted diagnosis. We report Differential Diagnosis Accuracy (DDx-ACC) and Critical Node Recall (CN-R). DDx-ACC is a common clinical measure for differential diagnosis. CN-R measures the proportion of essential DSM-guided nodes visited during the model’s reasoning trajectory. High CN-R reflects not only correct diagnostic labeling but also adherence to clinically appropriate diagnostic reasoning steps.

For the Tier~2 user experience evaluation, real users interacted with WiseMind and rated WiseMind’s Helpfulness and Empathy, two dimensions critical for patient engagement in psychiatric contexts. Following validated communication frameworks \cite{king2013best}, Helpfulness (Help.) reflects the degree to which responses offer guidance, clarify concerns, and provide relevant feedback, while Empathy (Emp.) captures perceived understanding, validation, and emotional attunement.

Six evaluators with backgrounds in human–computer interaction (HCI) and computing interacted with WiseMind as if conducting an intake conversation. Each evaluator was assigned a VSP biography but instructed to retell and elaborate it freely in their own words, introducing realistic conversational variability such as paraphrasing, omissions, narrative drift, emotional inflection, and spontaneous additions. This procedure mirrors standardized-patient training in clinical education and captures human-like noise that scripted dialogues cannot reproduce. All transcripts were anonymized prior to clinician review. This procedure resulted in 180 sessions (Real Human Interaction Corpus). After each interaction, the six evaluators rated WiseMind on helpfulness and empathy using a five-point Likert scale adapted from established clinical communication instruments~\cite{king2013best}. Scores were normalized to a 0–1 range. Full questionnaire items appear in Supplementary Section 6.2.

The Tier~3 doctor evaluation measured the clinical proficiency. To assess clinical credibility, licensed medical professionals evaluated WiseMind’s diagnostic reasoning, terminology, and communication style. Clinicians rated outputs along two Objective Structured Clinical Examination (OSCE)-derived criteria: Specialty (Spec.), the appropriateness of psychiatric terminology and reasoning patterns, and Precision (Prec.), the specificity and coherence of the diagnostic explanations \cite{dacre2003mrcp}.

Three mental-health professionals with a combined 33 years of clinical experience reviewed the Simulated Interaction Corpus and Real Human Interaction Corpus sampled from Tier~1 and Tier~2 sessions. Using 5-point Likert scales adapted from OSCE structured evaluation questionnaires, clinicians assessed whether WiseMind’s reasoning resembled novice-level clinical inquiry, adhered to DSM-guided diagnostic structure, and avoided clinically unsafe or misleading content. The final scores were normalized into 0-1. Full questionaire items appear in Supplementary Section 6.3.

We additionally conducted adversarial safety testing that included high-risk language (e.g., suicidality), contradictory symptom reports, and atypical conversational patterns. WiseMind’s escalation and repair mechanisms were evaluated in parallel with its diagnostic agents.

Across the three evaluation tiers, we assess WiseMind with simulated patients, real users, and licensed medical professionals. Table~\ref{tab:dataset-overview} summarizes the datasets supporting our diagnostic-, user-, and clinician-centered assessments. Ethical approvals and oversight procedures are described in the Methods section, and additional details on dataset construction and human evaluators are provided in Supplementary Section 5 and 6.

\begin{table*}[ht!]
\centering
\caption{Performance of WiseMind Architectural Components (GPT-4o base model), Evaluating Agent Structure, Knowledge Integration, and Action Spaces across Tier 1 Diagnostic Evaluation. \textbf{Tier 1 (Simulated Interaction Evaluation)} assesses agent structure, knowledge integration, and action spaces using synthetic interactions (e.g., virtual standardized patients and simulated dialogues) in controlled, repeatable settings.
\textbf{Knowledge Schemes}: KFP (Knowledge-Free Prompting); TKEP (Textual Knowledge-Enhanced: ICL for In-Context Learning, RAG for Retrieval-Augmented Generation); SKEP (Structured Knowledge-Enhanced via SKG - Structured Knowledge Graph).
\textbf{Agent Actions}: MC (Met Criteria); NMC (Not Met Criteria); MD (More Details); Contradict. (Contradiction).
\textbf{Metrics}: DDx (Differential Diagnosis Accuracy); CN-R (Critical Node Recall)
WiseMind variants: `Same-based' indicates a single LLM type for internal agents; `Mixed-based' uses different LLMs.
}
\label{tab:architectural_components}
\resizebox{\textwidth}{!}{%
\begin{tabular}{l c c c c}
\toprule
\multirow{2}{*}{\textbf{System Architecture}} & \multirow{2}{*}{\textbf{Knowledge Structure}} & \multirow{2}{*}{\textbf{Action Space}} & \multicolumn{2}{c}{\textbf{Tier 1: Diagnostic}} \\
\cmidrule(lr){4-5}
& & & DDx & CN-R \\
\midrule
\multicolumn{5}{l}{Benchmarking Systems} \\
\midrule
Random Baseline & None & - & 0.047  & - \\
Single Agent & None (KFP) & - & 0.316** ($\pm 0.13$) & - \\
Single Agent & Text (TKEP-ICL) & - & 0.356** ($\pm 0.11$) & 0.568** ($\pm 0.06$) \\
Single Agent & Text (TKEP-RAG) & - & 0.377** ($\pm 0.10$) & 0.422** ($\pm 0.07$) \\
Single Agent & Structural Graph (SKEP) & - & 0.707*~~ ($\pm 0.11$) & 0.888** ($\pm 0.07$) \\
\midrule
\multicolumn{5}{l}{Ablation Studies of WiseMind Components} \\
\midrule
WiseMind w/o SKG & TKEP & \{MC, NMC, MD, Contradict.\} & 0.386** ($\pm 0.11$) & 0.416** ($\pm 0.07$) \\
WiseMind w/o Multi-Agents & SKG & - & 0.707*~~ ($\pm 0.11$) & 0.888** ($\pm 0.08$) \\
WiseMind w/o MD & SKG & \{MC, NMC, Contradict.\} & 0.555** ($\pm 0.12$) & 0.643** ($\pm 0.09$) \\
WiseMind w/o Contradict. & SKG  & \{MC, NMC, MD\} & 0.498** ($\pm 0.12$) & 0.748** ($\pm 0.07$) \\
\midrule
\multicolumn{5}{l}{Complete WiseMind System} \\
\midrule
\textbf{WiseMind (Same-based)} & SKG & \{MC, NMC, MD, Contradict.\} & 0.802~~~~ ($\pm 0.10$) & 0.875~~~~ ($\pm 0.05$) \\
\textbf{WiseMind (Mixed-based)} & SKG & \{MC, NMC, MD, Contradict.\} & \textbf{0.856}~~~~ ($\pm 0.09$) & \textbf{0.956}~~~~ ($\pm 0.04$) \\
\midrule
Human Clinicians & Expert Knowledge & Clinical Interview & \textbf{0.860} \citep{regier2013dsm} & - \\
\bottomrule
\end{tabular}%
}

\begin{flushleft}
\linespread{1.0}\selectfont
\footnotesize{
\textbf{Statistical significance:}
Results are reported as Mean (95\% CI). The 95\% confidence intervals (CIs) for diagnostic accuracy (DDx) are calculated using the Wilson score interval, while CIs for other metrics (CN-R) are derived via 1000-iteration bootstrap resampling. Statistical significance between the best WiseMind systems (mixed-based) and benchmarking systems is annotated with asterisks (* $p < 0.05$, ** $p < 0.01$). The $p$-values are computed using McNemar's test for binary diagnostic outcomes (DDx) and the Wilcoxon signed-rank test for continuous metrics.}
\end{flushleft}
\end{table*}

\begin{table*}[ht!]
\centering
\caption{Performance of WiseMind Architectural Components (GPT-4o base model), evaluating Agent Structure, Knowledge Integration, and Action Spaces across Tier 2 and Tier 3 Human Evaluations.
\textbf{Tier 2 (Real Interaction Evaluation)} evaluates WiseMind on real human-interaction data to capture naturalistic conversational variability, with results reported separately from synthetic performance.
\textbf{Tier 3 (Cross-Setting Evaluation)} aggregates metrics across simulated and real interaction datasets to examine architectural robustness and consistency across controlled and real-world conditions. Results shown for average for different disorders. 
\textbf{Knowledge Schemes}: KFP (Knowledge-Free Prompting); TKEP (Textual Knowledge-Enhanced: ICL for In-Context Learning, RAG for Retrieval-Augmented Generation); SKEP (Structured Knowledge-Enhanced via SKG - Structured Knowledge Graph).
\textbf{Agent Actions}: MC (Met Criteria); NMC (Not Met Criteria); MD (More Details); Contradict. (Contradiction).
\textbf{Metrics}: Help. (Helpfulness); Emp. (Empathy); Spec. (Specialty); Prec. (Precision).
WiseMind variants: `Same-based' indicates a single LLM type for internal agents; `Mixed-based' uses different LLMs.
}
\label{tab:architectural_components_human_eval}
\resizebox{\textwidth}{!}{%
\begin{tabular}{lcccccc}
\toprule
\multirow{2}{*}{\textbf{System Architecture}} & \multirow{2}{*}{\textbf{Knowledge Structure}} & \multirow{2}{*}{\textbf{Action Space}}  & \multicolumn{2}{c}{\textbf{Tier2: User Exp}} & \multicolumn{2}{c}{\textbf{Tier~3: Doctor Eval}} \\
\cmidrule(lr){4-5} \cmidrule(lr){6-7}\\
& & &  Help. & Emp. & Spec. & Prec. \\
\midrule
\multicolumn{7}{l}{Benchmarking Systems} \\
\midrule
Single Agent & None (KFP)  & - & 0.575 & 0.417 & 0.418 & 0.466 \\
Single Agent & Text (TKEP-ICL)  & - & 0.447 & 0.275 & 0.477 & 0.625 \\
Single Agent & Text (TKEP-RAG)  & - & 0.548 & 0.421 & 0.437 & 0.525 \\
Single Agent & Structural Graph (SKEP)  & - & 0.584 & 0.513 & 0.615 & 0.648 \\
\midrule
\multicolumn{7}{l}{Ablation Studies of WiseMind Components} \\
\midrule
WiseMind w/o SKG & TKEP & \{MC, NMC, MD, Contradict.\}  & 0.602 & 0.675 & 0.529 & 0.551 \\
WiseMind w/o Multi-Agents & SKG  & - & 0.584 & 0.513 & 0.615 & 0.648 \\
WiseMind w/o MD & SKG & \{MC, NMC, Contradict.\}  & 0.453 & 0.398 & 0.592 & 0.649 \\
WiseMind w/o Contradict. & SKG  & \{MC, NMC, MD\}  & 0.692 & 0.685 & 0.537 & 0.492 \\
\midrule
\multicolumn{7}{l}{Complete WiseMind System} \\
\midrule
\textbf{WiseMind (Same-based)} & SKG & \{MC, NMC, MD, Contradict.\} & 0.701 & 0.692 & 0.673 & 0.673 \\
\textbf{WiseMind (Mixed-based)} & SKG & \{MC, NMC, MD, Contradict.\} &\textbf{0.825}&\textbf{0.816}&\textbf{0.697}&\textbf{0.768 }\\
\bottomrule
\end{tabular}%
}
\begin{flushleft}
\linespread{1.0}\selectfont
\footnotesize{
\textbf{Statistical significance:}
\textit{Note:} Results are reported as Mean. Tier-level inter-rater reliability (ICC2k) demonstrated excellent agreement for Tier 2 User Experience (0.96, 95\% CI: 0.92--0.98) and good agreement for Tier 3 Doctor Evaluation (0.75, 95\% CI: 0.42--0.92).}
\end{flushleft}
\end{table*}

\subsection*{Benchmarking Analysis}

We benchmark WiseMind against four representative prompting strategies that characterize current LLM-based approaches to diagnostic reasoning. Knowledge-Free Prompting (KFP) relies solely on naïve zero-shot prompting and pretrained priors. TKEP-ICL (Textual Knowledge--Enhanced Prompting with In-Context Learning) enriches prompts with exemplar diagnostic cases but still constrains the model to shallow pattern-matching. TKEP-RAG (Textual Knowledge--Enhanced Prompting with RAG) incorporates externally retrieved clinical text to improve factual grounding but lacks explicit reasoning scaffolds. The strongest single-agent baseline, SKEP (Structural Knowledge--Enhanced Prompting), integrates a SKG—a DSM-5–adapted decision graph encoding diagnostic entities and rule-based transitions—to impose a hierarchical differential-diagnosis trajectory. Together, these baselines span the range of contemporary prompting strategies from unstructured zero-shot prompting to structured knowledge-guided inference.

In Tier~1 simulated interaction evaluation, we evaluated the diagnostic performance of WiseMind. As shown in the Benchmarking results in Table~\ref{tab:architectural_components}, WiseMind substantially outperforms all baselines in diagnostic accuracy and structured reasoning. Averaged across depressive, bipolar, and anxiety disorders, the Mixed-based configuration achieves 0.856 DDx-ACC and 0.956 CN-R, while the Same-based configuration reaches 0.802 DDx-ACC and 0.875 CN-R. These performance gains over the strongest single-agent baseline (SKEP) are highly significant ($p < 0.01$ ; Odds Ratio = 3.0 -- 73.0, Cohen's g $\geq$ 0.25; McNemar's test for DDx; Wilcoxon signed-rank test for CN-R). In contrast, single-agent prompting methods—including the SKG-augmented SKEP—remain limited by reactive, single-pass generation that prevents proactive questioning, contradiction resolution, or integration of symptom evidence across turns. High CN-R scores show that WiseMind not only predicts correct diagnoses but also follows the appropriate DSM-guided diagnostic chains, reflecting expert clinical reasoning. In addition, WiseMind’s diagnostic accuracy is approaching the typical performance range of human clinicians (0.860 DDx) across psychiatric assessments \citep{basco2000methods,hirschfeld2003perceptions,shabani2021psychometric,osorio2019clinical}. These results are based on simulated interaction evaluations, which offer controlled testing conditions but do not reflect the full complexity of real-world psychiatric assessment. Accordingly, WiseMind is intended solely as an assistive tool that supports, rather than replaces, clinical judgment and must operate under human oversight in any real-world setting.

In the Tier~2 real interaction evaluation, we targeted the user experience such as helpfulness and empathy. To validate the consistency of these subjective evaluations, ratings from 6 user raters were aggregated, demonstrating excellent inter-rater reliability (ICC2k = 0.96; 95\% CI: 0.92--0.98; 6 raters). Benchmarking results in Table~\ref{tab:architectural_components_human_eval} show that WiseMind produces markedly more supportive and empathic interactions than all single-agent baselines. KFP and TKEP-ICL generate largely generic supportive statements (Help. = 0.447--0.575; Emp. = 0.275--0.417). TKEP-RAG improves factual grounding but remains emotionally shallow, while SKEP shows improved coherence (Help. = 0.584; Emp. = 0.513). In contrast, WiseMind’s Mixed-based configuration achieves the highest ratings (Help. = 0.825; Emp. = 0.816). This indicates that separating structured reasoning (Reasonable Mind) from empathic reformulation (Emotional Mind) produces interactions that evaluators consistently rated as clearer, more validating, and more clinically appropriate. The Same-based model (Help. = 0.701; Emp.= 0.692) also surpasses all baselines, though it remains slightly below the Mixed-based design, suggesting that heterogeneous agent specialization deepens empathic attunement.

Finally, in Tier~3 simulated and real interaction evaluation, a panel of 3 medical experts conducted evaluations to measure the specialty and precision of the WiseMind system. Despite the inherent subjectivity of individual clinical judgment, the expert panel demonstrated good tier-level inter-rater reliability (ICC2k = 0.75; 95\% CI: 0.42--0.92; 3 raters), confirming the stability of the aggregated clinical consensus. Results in Table~\ref{tab:architectural_components_human_eval} confirm that WiseMind achieves the strongest clinician-rated specialty (Spec.) and precision (Prec.). Baseline prompting methods receive low Spec. (0.418--0.477) and Prec. (0.466--0.625), with clinicians noting inconsistent psychiatric terminology and unstable reasoning patterns across turns. SKEP performs better (Spec. = 0.615; Prec. = 0.648), but explanations often lacked an overarching diagnostic strategy. The WiseMind Same-based variant surpasses all baselines (Spec. = 0.673; Prec. = 0.673), while the Mixed-based configuration achieves the highest scores (Spec. = 0.697; Prec. = 0.768). Clinicians emphasized that WiseMind adhered closely to DSM-guided inquiry patterns, used appropriate psychiatric vocabulary, and demonstrated early-career–level reasoning fidelity.

\subsection*{Ablation Analysis}

To isolate the contribution of each architectural component, we conduct a suite of ablation experiments in which structural and functional elements of WiseMind are selectively removed. The w/o SKG variant eliminates the Structured Knowledge Graph, forcing reliance on unstructured retrieval. The w/o Multi-Agents variant merges the Emotional Mind and Reasonable Mind into a single-agent system, removing the separation between empathic interviewing and structured reasoning. We also ablate elements of the action space by removing the More Details (MD) action—responsible for refining ambiguous symptom descriptions—and the Contradiction action, which identifies and repairs inconsistencies across turns.

For Tier~1, the ablation analysis panel in Table~\ref{tab:architectural_components} shows that removing SKG yields a dramatic collapse in diagnostic reliability: DDx-ACC drops to 0.386 and CN-R to 0.416 ($p < 0.01$). Removing the Multi-Agent architecture reduces DDx-ACC to 0.707 ($p < 0.05$), confirming that separating empathic and analytical functions is essential for maintaining diagnostic rigor. Removing the MD action reduces performance to 0.555 ($p < 0.01$), indicating that proactive clarification is necessary for resolving uncertain or partially expressed symptoms. Eliminating Contradiction detection leads to major error propagation and drops the diagnostic performance to 0.498 ($p < 0.01$), showing that inconsistency management is critical for safe psychiatric assessment. Across variants, removing any single component reduces diagnostic performance by 15--47 points ($p < 0.05$), demonstrating that WiseMind’s performance arises from synergistic design rather than any isolated mechanism.

For Tier~2, Table~\ref{tab:architectural_components_human_eval} shows that removing structural elements disrupts user experience quality. Without the SKG, Help. and Emp. fall to 0.602 / 0.675, indicating that clinical structure supports more relevant and context-sensitive responses. Removing Multi-Agents reduces Emp. to 0.513, demonstrating the unique role of the Emotional Mind agent in maintaining conversational warmth. Removing the MD action yields the steepest decline in user experience (Help. = 0.453; Emp. = 0.398), as the system fails to clarify ambiguous user statements—an essential component of empathic interviewing. Removing contradiction detection reduces emotional coherence and disrupts conversational flow, lowering Emp. to 0.685. These results demonstrate that WiseMind’s supportive interactions depend on the interplay of SKG-driven structure, agent specialization, and action-level behaviors.

Finally for Tier~3, As shown in Table~\ref{tab:architectural_components_human_eval}, clinicians rated ablated models substantially lower. Without the SKG, Spec. and Prec. drop to 0.529 / 0.551, with clinicians noting loss of structured psychiatric scaffolding. Removing Multi-Agents reduces scores to SKEP-level performance (0.615 / 0.648), showing that the two-agent architecture is crucial for professional tone and reasoning fidelity. Removing the MD action lowers Spec. and Prec. to 0.592 / 0.649, as the system struggles to refine ambiguous patient statements. Removing contradiction detection produces the lowest scores across all ablations (0.537 / 0.492), as the system fails to correct inconsistencies—an essential competency in clinical assessment. These results show that WiseMind’s clinical credibility emerges from the joint contribution of structured knowledge, agent specialization, and action-space design.

\subsection*{Qualitative Feedback from Clinicians}

During the Tier~3 Doctor Evaluation, Clinician also provided qualitative feedback, highlighting multiple dimensions of WiseMind’s performance as an assistive conversational tool. In terms of helpfulness, participants noted that “WiseMind asked the right follow-up questions at the right time. It felt like the system was actively helping me clarify what I wanted to express,” and that “the interaction gave me meaningful guidance instead of generic responses—far better than standard chatbots.” WiseMind was also consistently described as empathic, with clinicians reporting that “the agent consistently responded in a warm and understanding tone. It made the experience feel human rather than mechanical,” and that “WiseMind validated my feelings without overdoing it. It struck a good balance between emotional support and professionalism.” Under specialty, reviewers emphasized that “the terminology WiseMind used was clinically appropriate—neither too technical nor too simplified,” and that “it followed a diagnostic structure that reflects real psychiatric intake interviews. The flow felt medically familiar.” Finally, WiseMind was praised for its precision: “Its reasoning was coherent and specific. Each question had a clear purpose tied to the diagnosis,” and “WiseMind avoided unnecessary detours. It focused on details that actually matter in differential diagnosis.” Sample comments are included in Supplementary Section 8. Importantly, the interactions evaluated include both simulated and real interaction settings. Because real-world psychiatric assessment is substantially more complex, WiseMind is intended to function as an assistive tool under human oversight.


\begin{figure*}[ht!]
\centering
\includegraphics[width=\linewidth]{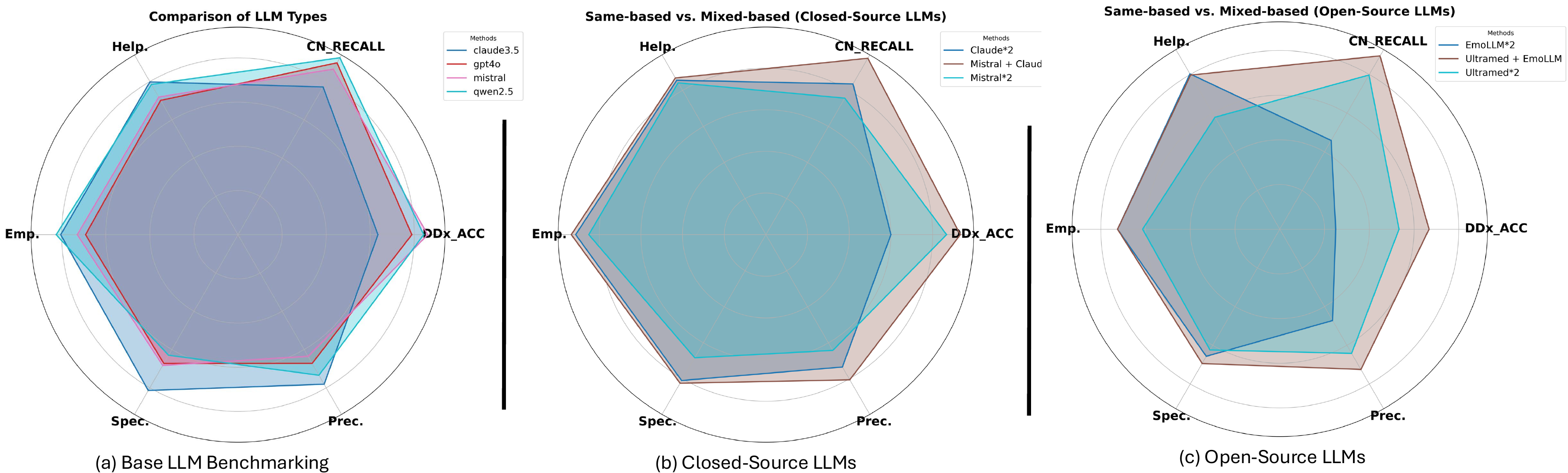}
\caption{Comparative performance of the WiseMind framework across base models and multi-agent configurations. (a) Performance comparison across different LLMs, illustrating that individual models exhibit distinct strengths across the WiseMind’s reasonable-mind and emotional-mind dimensions. (b) Performance comparison of same-base versus mixed-base multi-agent configurations using closed-source LLMs. (c) Corresponding comparison for open-source LLMs. Across both closed- and open-source settings, assigning different LLMs to the Reasonable Mind Agent (RA) and the Emotional Mind Agent (EA)—based on their task-specific strengths—consistently improves diagnostic accuracy, empathy, and clinical realism compared with using the same model for both roles. Differential Diagnosis Accuracy (DDx-ACC) and Critical Node Recall (CN-R) are computed using simulated interactions; Helpfulness (Help.) and Empathy (Emp.) are evaluated using real interactions, and Specialty (Spec.) and Precision (Prec.) are assessed across both simulated and real interactions.}
\label{fig:performance_mix}
\end{figure*}

\subsection*{Comparison Across LLM Model Selection}
\label{sec:comparativeLLM}

In this section, we compare the effects of different open-source and closed-source LLMs on overall system performance across all three tiers, as shown in Fig.~\ref{fig:performance_mix}. Detailed performance results are provided in Supplementary Section 7 and Supplementary Table 6.  We first systematically benchmarked four families of widely used models—GPT-4o, Claude~3.5, 
Mistral-7B, and Qwen2.5—chosen to represent diverse training paradigms, license types, 
and performance characteristics. GPT-4o and Claude~3.5 exemplify state-of-the-art 
closed-source models known for strong reasoning and conversation stability, whereas 
Mistral-7B and Qwen2.5 represent high-performance open-source models with practical 
deployment advantages. This selection reflects the models most frequently used in 
clinical NLP and psychiatric simulation research.

Fig.~\ref{fig:performance_mix}a reveals that each model exhibits distinct performance 
profiles, demonstrating clear specialization patterns. Some models, such as Claude~3.5, 
excel in generating responses rated highly for empathy and clinical specialty, making 
them naturally well-suited for patient-facing interaction. Others, such as Qwen2.5 and 
Mistral, demonstrate stronger diagnostic performance and symptom coverage (high 
CN-R) while yielding more moderate user experience scores. GPT-4o exhibits 
particularly strong structural reasoning and node coverage but shows variability across 
interaction quality metrics. 

These benchmarking results indicate that LLM backbones possess intrinsic predispositions—
some optimized toward analytical reasoning and others toward conversational empathy—due 
to differences in pretraining corpora and alignment objectives. When a single model is 
used to serve both WiseMind agents (RA and EA), 
its weaknesses inevitably constrain overall system performance. For instance, a 
reasoning-strong model may provide coherent diagnostic sequences but struggle to convey 
emotional attunement, while an empathy-oriented model may excel at rapport-building yet 
produce less precise clinical terminology. These trade-offs observed across 
individual base models further reinforce that heterogeneity in agent design 
may be essential for achieving balanced performance.

Motivated by this observation, we introduced mixed-base configurations that 
assign different LLMs to RA and EA according to their strengths. As shown in 
Figures~\ref{fig:performance_mix}b and \ref{fig:performance_mix}c, this strategy 
consistently yields substantial performance gains across all evaluation dimensions. 
For closed-source models, combinations such as Mistral (RA) + Claude (EA) achieve 
improvements in diagnostic accuracy, empathy, and specialty. Open-source pairings 
show a similar pattern: using UltraMed for RA alongside EmoLLM for EA—both derived from 
LLaMA3-8B—results in markedly stronger performance than using either model alone. These 
findings underscore a general design principle: role–model alignment is critical 
for maximizing both reasoning quality and interactional sensitivity. Mixed-base 
architectures allow WiseMind to capitalize on complementary strengths rather than 
forcing a single model to balance incompatible demands.

\subsection*{Comparison Across Psychiatric Disorders}
\label{sec:comparativeOrder}
Table \ref{tab:tier1_benchmark} presents WiseMind's performance on simulated interaction evaluations across depression, bipolar disorder, and anxiety. The breakdown reveals distinct patterns across disorders. For depression, SKEP with single agent already offers substantial gains over earlier prompting approaches, but both dual-agent WiseMind variants improve further, with the mixed-base configuration achieving the highest DDx (0.875). Bipolar disorder—often more challenging due to overlapping mood symptoms—shows the clearest separation between approaches: SKEP provides a strong single-agent baseline, yet WiseMind’s mixed-base configuration markedly improves both DDx (0.933) and CN-R (0.978), suggesting enhanced discrimination of complex symptom trajectories. Anxiety displays the highest CN-R across conditions; both mixed- and same-base WiseMind variants maintain strong DDx while achieving near-ceiling recall, indicating stable detection of anxiety-related symptom cues.

Across all three disorders, WiseMind’s mixed-base diagnostic accuracy on simulated interaction evaluations is approaching the published performance ranges of human clinicians. However, real-world psychiatric assessment involves greater complexity, and these results do not imply clinical equivalence or replacement of human clinicians. Regardless, his alignment underscores the clinical relevance of WiseMind’s role-aligned architecture, particularly in diagnostically complex categories such as bipolar disorder. Notably, similar patterns hold in Tier~2 and Tier~3 evaluations (helpfulness, empathy, specificity, and precision), where mixed-base configurations again outperform same-base and single-agent systems. This convergence across diagnostic and affective metrics reinforces the value of combining complementary model strengths within the WiseMind framework.

\begin{table*}[ht]
\centering
\caption{Diagnostic performance across three mental disorders.}
\label{tab:tier1_benchmark}
\setlength{\tabcolsep}{2.5pt}

\resizebox{\textwidth}{!}{%
\begin{tabular}{lcccccc}
\toprule
\multirow{2}{*}{\textbf{System}} 
& \multicolumn{2}{c}{\textbf{Depression}} 
& \multicolumn{2}{c}{\textbf{Bipolar}} 
& \multicolumn{2}{c}{\textbf{Anxiety}} \\
\cmidrule(lr){2-3} 
\cmidrule(lr){4-5} 
\cmidrule(lr){6-7}
& DDx & CN-R 
& DDx & CN-R 
& DDx & CN-R \\
\midrule

Random Guess              & 0.040   & - & 0.063  & - & 0.038  & - \\
Single Agent (KFP)        & 0.256** ($\pm 0.18$) & - & 0.293** ($\pm 0.19$) & - & 0.400*~~ ($\pm 0.18$) & - \\
Single Agent (TKEP-ICL)   & 0.280** ($\pm 0.17$) & 0.466** ($\pm 0.10$) & 0.427** ($\pm 0.09$) & 0.721** ($\pm 0.11$) & 0.360** ($\pm 0.18$) & 0.517** ($\pm 0.09$) \\
Single Agent (TKEP-RAG)   & 0.250** ($\pm 0.15$) & 0.235** ($\pm 0.06$) & 0.400** ($\pm 0.22$) & 0.739** ($\pm 0.11$) & 0.480*~~ ($\pm 0.19$) & 0.292** ($\pm 0.08$) \\
Single Agent (SKEP)       & 0.555*~~ ($\pm 0.19$) & 0.800** ($\pm 0.13$) & 0.767*~~ ($\pm0.13$) & 0.891~~~~ ($\pm0.11$) & 0.800~~~~ ($\pm0.15$) & \textbf{0.974}~~~~ ($\pm0.03$) \\
\midrule

\textbf{WiseMind (Same-based) }
& 0.833~~~~ ($\pm 0.15$) & \textbf{0.956}~~~~ ($\pm 0.05$)
& 0.733~~~~ ($\pm 0.21$) & 0.769~~~~ ($\pm 0.07$) 
& \textbf{0.840}~~~~ ($\pm 0.15$) & 0.899~~~~ ($\pm 0.08$) \\

\textbf{WiseMind (Mixed-based)} 
& \textbf{0.875}~~~~ ($\pm0.14$) & 0.939~~~~ ($\pm0.08$) 
& \textbf{0.933~~~~ ($\pm0.15$) }& \textbf{0.978~~~~ ($\pm0.04$) }
& 0.760~~~~ ($\pm0.16$) & 0.951~~~~ ($\pm0.04$) \\
\midrule

Human Clinicians \cite{basco2000methods,hirschfeld2003perceptions,shabani2021psychometric,osorio2019clinical} 
& 0.700--0.900 & - 
& 0.600--0.800 & - 
& 0.700--0.850 & - \\
\bottomrule
\end{tabular}%
}
\begin{flushleft}
\linespread{1.0}\selectfont 
\footnotesize{
\textbf{Statistical significance:}
Results are reported as Mean (95\% CI). The 95\% confidence intervals (CIs) for diagnostic accuracy (DDx) are calculated using the Wilson score interval, while CIs for other metrics (CN-R) are derived via 1000-iteration bootstrap resampling. Statistical significance between the best WiseMind systems (Mixed-based) and benchmarking systems is annotated with asterisks (* $p < 0.05$, ** $p < 0.01$). The $p$-values are computed using McNemar's test for binary diagnostic outcomes (DDx) and the Wilcoxon signed-rank test for continuous metrics.}
\end{flushleft}
\end{table*}

\subsection*{Error Analysis and Risk Management Discussion}

To capture nuances of real-world clinical usage and evaluate downstream safety risks, we
implemented a structured adversarial testing and ethical risk–mitigation protocol.
WiseMind includes a dedicated Risk Management module that integrates escalation
logic for high-risk situations and automated repair mechanisms for recoverable errors. In
addition, the system dynamically adapts its interaction strategy in response to
conversational imbalances—such as minimal disclosure, contradictory symptom reports, or
excessively verbose input—following the behavioral guidelines detailed in Supplementary Section 4. This framework enables systematic assessment of WiseMind’s robustness under both
system-generated and user-generated sources of failure.

\begin{table}[!h]
\centering
\caption{Outcome breakdown of ethical test cases. A case is marked as resolved if the system internally corrected its behavior and produced a safe or correct diagnosis. Escalation indicates backend alert or a safety stop.}
\label{tab:adv_breakdown}
\scriptsize

\begin{tabular}{lccc}
\toprule
\textbf{Category} & \textbf{Cases} & \textbf{Resolved} & \textbf{Escalated} \\
\midrule

\multicolumn{4}{l}{\textbf{Intrinsic System Errors}} \\
RA Decision Errors            & 5 & 3 & 1 \\
EA Generation Errors          & 5 & 4 & 2 \\
\midrule

\multicolumn{4}{l}{\textbf{Extrinsic Adversarial Behaviors}} \\
Suicidal/Homicidal Talk       & 5 & 5 & 5 \\
Contradictions in Responses   & 5 & 4 & 1 \\
Under-talking (Minimal Input) & 5 & 3 & 0 \\
Over-talking (Verbose Input)  & 5 & 5 & 2 \\
\midrule

\textbf{Total}                & 30 & 24 & 11 \\
\bottomrule
\end{tabular}
\end{table}

To characterize WiseMind’s behavior under challenging conditions and evaluate its safety
controls, we conducted a 30-case adversarial stress test on simulated interactions between VSP and WiseMind, spanning both intrinsic system
failures and extrinsic adversarial user behaviors (Table~\ref{tab:adv_breakdown}). A case
was marked as resolved when the system internally corrected its behavior and
produced a safe or clinically appropriate diagnosis, and as escalated when it
triggered a backend alert or safety stop. A case can be resolved and escalated at the same time once the error potentially causes catastrophic consequences (i.e. risky response, misleading hallucinations which contradict the medical facts).

The intrinsic system errors capture failures originating from WiseMind’s own reasoning or generation
processes. We tested five cases involving RA decision errors and five cases involving EA
generation errors. As shown in Table~\ref{tab:adv_breakdown}, WiseMind successfully
resolved 3 of 5 RA failures and 4 of 5 EA failures, typically via structured recovery
behaviors such as revisiting missed SKG nodes, issuing clarification prompts, or
rephrasing questions to repair incoherent dialogue turns. These results suggest that the
dual-agent architecture, coupled with explicit fallback strategies, provides meaningful
self-correction for a majority of internal failures. At the same time, a subset of cases
(1 RA and 2 EA) triggered escalation.

The extrinsic adversarial  behaviors cover challenging patient communication patterns that can destabilize
psychiatric interviews even for human clinicians. WiseMind showed strong reliability in
handling these scenarios, as shown in Table~\ref{tab:adv_breakdown}. All five suicidal or homicidal-talk cases were both resolved
and escalated, indicating that the system consistently recognized high-risk language and
activated the safety protocol without missing any critical event. For contradictory
responses, 4 of 5 cases were resolved and 1 escalated, reflecting appropriate detection
of inconsistencies and return to key diagnostic branches. Conversational imbalances were
also handled adaptively: in under-talking (minimal-input) scenarios, WiseMind resolved 3
of 5 cases by shifting toward more guided questioning, whereas in over-talking
(verbose-input) scenarios it resolved all 5 cases and escalated 2, primarily through
summarization and refocusing strategies. 

Overall, WiseMind resolved 24 of 30 adversarial
cases and escalated 11, demonstrating that its layered safety logic can manage both
system-driven and user-driven risks while retaining conservative behavior in the highest
risk situations.

\section*{Discussion}

This work demonstrates that large language models can be recontextualized for psychiatric differential diagnosis by embedding them within structured knowledge, process-aware modeling, and holistic evaluation. Rather than treating psychiatric interviews as another language benchmark, our framework emphasizes that clinical utility depends on faithfully reproducing the decision logic of the DSM, balancing empathic and analytic roles during dialogue, and establishing safety and trustworthiness through comprehensive evaluation. The results highlight not only methodological advances but also broader lessons for how to move clinical AI closer to real-world adoption \citep{topol2019high,He2024Trustworthy}.

A central finding is that structured knowledge is indispensable for reliable diagnostic reasoning. Conventional methods that rely on unstructured text retrieval or KFP, such as standard RAG or ICL)\citep{dong-etal-2024-survey}, falter when faced with the combinatorial complexity of psychiatric differential diagnosis, where disorders overlap in nuanced and hierarchical ways \citep{mcduff2025towards,First2024DSM5TR}. By contrast, encoding DSM criteria as a state-transition graph enabled the system to proactively navigate diagnostic pathways, reflecting the formal structure of clinician-guided diagnostic workflows \citep{Demazeux2015DSM5}. This shift from reactive recall to proactive, decision-guided inquiry reflects a broader principle: in high-stakes medicine, AI must operate not just as a knowledge repository but as a process-aware participant that enforces clinical rigor \citep{Yan2022challenge,rajpurkar2022ai}.

Equally important is the demonstration that diagnostic accuracy and empathic engagement are not competing goals. Psychiatric interviews unfold as a delicate balance between analytic questioning and the establishment of a therapeutic alliance \citep{norcross2011working}. Our dual-agent design, inspired by DBT’s concept of the “reasonable mind” and the “emotional mind” \citep{linehan1993cognitive,Chapman2006DBTCurrentIndications}, provides empirical evidence that role separation enables both dimensions to flourish. Specifically, the RA consults the knowledge graph to guide structured diagnostic reasoning, while the EA reformulates actions in empathic, trust-building language. The observed improvements in empathy, trust, and rapport, without sacrificing diagnostic precision, suggest that multi-agent specialization may be a general design paradigm for medical AI. Particularly striking was the finding that mixed-base configurations, which assign different underlying LLM models to each role according to their natural strengths, achieved the best overall performance in both simulated and real interactions. This points to a future where modular orchestration of specialized AI agents, rather than reliance on monolithic models, becomes a cornerstone of clinically robust systems \citep{10.1145/3586183.3606763}.

Beyond technical advances, the design of WiseMind aligns with broader frameworks of medical professionalism. According to the CanMEDS framework \citep{Frank2015CanMEDS}, a competent physician is defined not only as a \emph{medical expert} but also as a \emph{communicator}, \emph{collaborator}, \emph{leader}, \emph{health advocate}, \emph{scholar}, and \emph{professional}. Yet much of the work in medical AI has focused narrowly on diagnostic accuracy or reasoning performance, often overlooking equally critical dimensions such as communication, ethical responsibility, and collaboration. WiseMind explicitly incorporates these dimensions. The separation of the Reasonable Mind and Emotional Mind agents reflects the dual role of doctors as both rigorous diagnosticians and empathetic communicators. The multi-agent architecture itself embodies the principle of collaboration, demonstrating how specialized roles can complement one another within a shared clinical process. Risk management protocols and ethical safeguards operationalize the role of the physician as a professional, responsible not only for accuracy but also for protecting privacy, managing risk, and ensuring patient safety. Our multi-faceted evaluation, which spans user trust and clinician review in addition to accuracy, parallels the CanMEDS vision that medical expertise must be judged in concert with communication, professionalism, and leadership. In this sense, the significance of WiseMind lies not only in its performance metrics but also in its embodiment of the broader competencies required for clinical practice. By embedding CanMEDS-inspired principles into system design, we move toward conversational diagnostic agents that are not just technically capable, but also aligned with the humanistic and professional standards that define real-world medicine.

Our evaluation strategy further underscores the importance of expanding beyond narrow technical metrics. In psychiatry, the success of an AI system cannot be judged by accuracy alone. The quality of patient experience, the preservation of therapeutic rapport, and the system’s ability to behave responsibly under adversarial or high-risk scenarios are equally vital \citep{moon2023ethical,kerz2023toward}. By incorporating simulated patients, human user studies, clinician assessments, and adversarial stress tests, we demonstrate that it is possible to measure these broader dimensions in a systematic way. The results suggest that when safety protocols, empathy, and reasoning transparency are treated as first-class objectives, AI systems can achieve a level of robustness and user-perceived appropriateness beyond what conventional benchmarks capture.

Taken together, these contributions indicate that psychiatric AI requires a shift in design philosophy: from maximizing model accuracy in isolation toward creating modular, theory-informed systems that integrate domain knowledge, clinical processes, and ethical safeguards. Such systems hold promise as triage assistants or early screening tools under human supervision, helping clinicians prioritize scarce time and resources while maintaining the therapeutic alliance that underpins effective psychiatric care, rather than fully replacing clinical judgment. More broadly, the principles demonstrated here may extend to other high-stakes fields such as oncology, pathology, or even education, where trust, empathy, and structured reasoning are prerequisites for real-world deployment \citep{davenport2019future}.

Ethical considerations remain central to the translation of this work. WiseMind is explicitly designed as an assistive technology to augment, not replace, clinical expertise. Human oversight must remain integral, and deployment must be accompanied by safeguards addressing bias, privacy, and transparency. The inclusion of escalation protocols for high-risk scenarios illustrates one approach, but continued refinement and regulatory alignment will be essential \citep{price2019privacy}. By embedding safety and humanistic values into the system’s very architecture, we aim to contribute not only a technical advance but also a paradigm for responsible psychiatric AI.

Nonetheless, important limitations remain. Our study focused on three disorders—depression, anxiety, and bipolar disorder—leaving open questions about scalability across the full spectrum of DSM diagnoses. In addition, the evaluations on real interactions between human and WiseMind were conducted with a relatively small number of participants (six users and three clinicians), which constrains statistical power and limits the generalizability of qualitative and quantitative findings. Although we employed hundreds of high-fidelity simulated patients and human evaluations, clinical trials with diverse real-world populations will be necessary to establish generalizability and longitudinal impact. Furthermore, diagnostic comparisons with clinicians were conducted under simulated interaction settings rather than prospective clinical trials. WiseMind is therefore intended as an assistive system requiring human oversight, and future large-scale, real-world validation will be essential before clinical integration. The current reliance on expert-curated knowledge graphs also raises challenges of scalability, future work may explore semi-automated updating methods and integration with alternative frameworks such as ICD-11 to enhance adaptability.  Moreover, while our adversarial testing showed encouraging robustness, systematic evaluation across cultural, linguistic, and demographic variations will be critical for ensuring fairness and equity \citep{pierson2021demographic}, especially since the framework was tested primarily in English and with simulated patient profiles that may not reflect non-Western symptom expressions or socioeconomic diversity. Finally, reliance on DSM-5 criteria reflects a Western nosology that may not capture cultural variation in psychiatric presentations, underscoring the need for global perspectives and cross-cultural validation.

In conclusion, we present WiseMind, a clinically aligned multi-agent system designed to support psychiatric assessment under human oversight. We address key limitations of current medical AI systems in structured reasoning, adaptive interviewing, and empathic communication by integrating a DSM-5–guided SKG with a DBT–informed dual-agent architecture. The dual-agent architecture combines an RA for criterion-based diagnostic reasoning and an EA for supportive, patient-centered dialogue. Using VSP, real-user interactions, and expert clinician evaluations, we demonstrate that WiseMind substantially improves general LLM performance across differential diagnosis accuracy, critical diagnostic-path traversal, conversational empathy, and safety. These results indicate that WiseMind offers a feasible and transparent framework for building trustworthy psychiatric AI, and we hope this work can catalyzes further research toward clinically coherent, emotionally attuned, and risk-aware LLM systems in digital mental health.

\section*{Methods}

\subsection*{Structured Knowledge Graph (SKG)}
In professional counseling sessions—particularly in the context of mental disorder diagnosis—the conversation is not an open-ended exchange but a structured, dynamic clinical interview \citep{Sommers-Flanagan2023}. A diagnostic agent must therefore \emph{proactively} guide the dialogue to ensure that critical criteria are systematically assessed, rather than merely reacting to patient-initiated questions. This requirement implies the need for an explicit agenda or roadmap that allows the agent to navigate the diagnostic process in a coherent and clinically valid manner. To meet this fundamental requirement, we implemented an SKG derived from the DSM-5 decision pathways, which serves as a guideline for the LLM. The knowledge graph provides the agent with both the sequence and logic of diagnostic steps, enabling it to conduct interviews that are comprehensive, consistent, and aligned with professional psychiatric practice.

 DSM-5 differential-diagnosis decision pathways \citep{First2024DSM5TR} were encoded as a \emph{structured knowledge graph} $G$ with nodes $v \in G$ representing diagnostic criteria or decision points. Nodes store criterion details $c_i$ and decision logic $d_i$ (e.g., temporal constraints, exclusion rules), and edges implement state transitions along DSM-style pathways (hierarchical symptom clusters, specifiers, temporal patterns). This enables structured, proactive navigation rather than reactive recall while keeping strong alignment to the well-established diagnostic instruction (DSM-5).

\begin{figure*}[!h]
    \centering
    \includegraphics[width=0.7\linewidth]{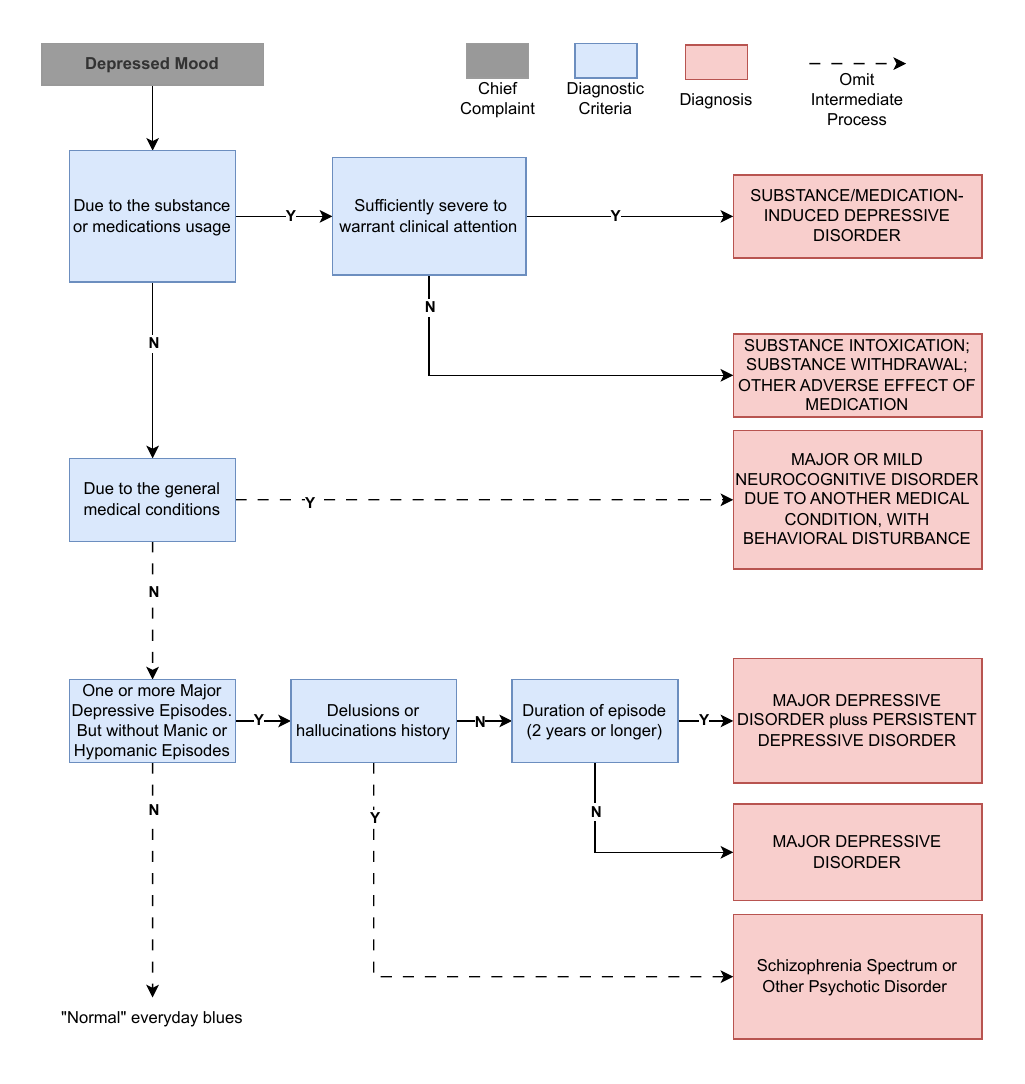}
    
    \caption{Example of the DDx decision tree for depressed mood. Visualization of the DSM-5-TR differential diagnosis decision tree that served as the source for constructing the structured knowledge graph. The flowchart outlines the key assessment topics, exclusion rules, and diagnostic pathways used to differentiate causes of depressed mood.}
    \label{fig:dsm-ddx}
\end{figure*}

 During an interview, the SKG provides a \emph{reasoning scaffold}: given dialogue history $D$, the system retrieves context-aware knowledge $K = R(v, D, G)$, selects an action $a = P(K, r, D)$ contingent on the latest patient response $r$, and transitions $v \leftarrow T(v, a, G)$ to advance or branch the DDx path. This mirrors clinical interviews that follow DSM-guided decision points to elicit missing evidence and rule out confounds, effectively using the knowledge graph as a task-oriented agenda \citep{edge2024local}. 

 An example of the original DDx decision tree for depressed mood is shown in Fig.~\ref{fig:dsm-ddx}. This decision tree defines a common procedure for a clinician to conduct DDx when interviewing patients. It defines the critical topics that must be assessed and their main criteria. The structured knowledge graph is modified from these types of decision trees.

\subsection*{Dual-Agent Architecture}
Inspired by DBT theory \citep{Chapman2006DBTCurrentIndications}, WiseMind decomposes psychiatric assessment into two specialized LLM agents—a RA and an EA—each with distinct but complementary functions. Both agents access the shared SKG (denoted as $G$) and dialogue history ($D$), but utilize them for different purposes in the diagnostic process.

The RA serves as the cognitive engine of WiseMind. Its primary responsibility is diagnostic decision-making based on the structured clinical knowledge encoded in the SKG. The RA processes patient responses against current diagnostic criteria and determines the appropriate next action from a discrete action space: $met\_criteria$ (criterion satisfied), $not\_met\_criteria$ (criterion not satisfied), $needs\_more\_information$ (clarification required), or $contradiction$ (response contradicts previous information).

The RA's decision process follows a systematic two-step reasoning workflow. First, it retrieves relevant diagnostic knowledge by interpreting dialogue history $D$ in the context of the knowledge graph $G$: $K_t = R(v_t, D, G)$, where $R(\cdot)$ is the retrieval function that maps the current node state $v_t$ and dialogue history to pertinent knowledge from the SKG. Second, it determines the appropriate action: $a_t = P(K_t, r_t, D)$, where $P(\cdot)$ is the policy function that selects a diagnostic action based on the retrieved knowledge $K_t$ and the patient's latest response $r_t$. Based on this action, the RA transitions to the next relevant node in the knowledge graph: $v_{t+1} = T(v_t, a_t, G)$, where $T(\cdot)$ is the state transition function defined by the structural relationships in the SKG.

The EA functions as the therapeutic communicator of the system, focusing on building rapport. While the RA determines what information to gather, the EA determines how to elicit it empathically. The EA formulates natural language that achieves multiple therapeutic goals: eliciting necessary diagnostic information, maintaining appropriate emotional tone, building rapport, and adapting to the patient's unique communication style.

Formally, the EA computes the next utterance $q_{t+1}$ as: $q_{t+1} = Q(v_{t+1}, a_t, D, G)$, where $Q(\cdot)$ is the generation function that considers the diagnostic target node $v_{t+1}$ (selected by the RA), the RA's action $a_t$, dialogue history $D$, and relevant content from the SKG $G$ to construct a contextually appropriate and empathic question.

The synergy between these agents—each optimized for its specialized function but sharing access to common knowledge sources—creates a unified diagnostic experience that balances systematic assessment with emotional attunement. Their coordinated workflow, formalized in Algorithm~\ref{alg:wisemind_workflow}, enables WiseMind to conduct clinically rigorous yet emotionally responsive conversations that more closely approximate human psychiatric practice.

\begin{algorithm}
\caption{WiseMind diagnostic loop (condensed)}
\label{alg:wisemind_workflow}
\begin{algorithmic}[1]
\State Input: initial complaint $C_{\text{init}}$, SKG $G$;\quad Init: $D \!\leftarrow\! [\mathrm{Greet}(EA),\, C_{\text{init}}]$, $v \!\leftarrow\! \mathrm{GetStartNode}(G, C_{\text{init}})$
\While{not $\mathrm{Terminated}(v, D)$}
    \State $r \!\leftarrow\! \mathrm{LastPatientResponse}(D)$
    \State $K \!\leftarrow\! R(v, D, G)$ \Comment{retrieve DSM-guided context}
    \State $a \!\leftarrow\! P(K, r, D)$ \Comment{RA action: met, not met, MD, contradiction}
    \State $v \!\leftarrow\! T(v, a, G)$ \Comment{advance along DSM pathway}
    \State $q \!\leftarrow\! Q(v, a, D, G)$ \Comment{EA generates empathic question}
    \State append $q$ to $D$; present to patient; receive $r'$; append to $D$
\EndWhile
\State Output: final diagnostic state inferred from $v$ and $D$
\end{algorithmic}
\end{algorithm}

\subsection*{Model Implementation}

The computational setup used in all experiments is summarized in 
Table~\ref{tab:model-implementation}, including the model checkpoints evaluated, 
hardware configuration, inference settings, runtime considerations, and overall 
computational cost. All models were run either through their respective APIs or 
locally on our multi-GPU workstation without additional finetuning.

\begin{table*}[!h]
\centering
\caption{Model Implementation Overview. Summary of all LLM checkpoints, 
hardware infrastructure, software environment, and inference configurations used 
in the WiseMind framework. Closed-source models were accessed via API, whereas 
open-source models were deployed locally using vLLM.}
\label{tab:model-implementation}
\small
\setlength{\tabcolsep}{8pt}
\begin{tabular}{p{3.5cm} p{8.5cm}}
\toprule
\textbf{Component} & \textbf{Description} \\
\midrule

\textbf{LLM Checkpoints }
& GPT-4o (OpenAI); Claude-3.5-Sonnet (Anthropic); 
Mistral-Large-Latest (Mistral AI API); Qwen2.5-72B (Ollama); 
EmoLLM (LLaMA3-8B, emotional variant); 
UltraMedical (LLaMA3-8B, medical variant). 
All checkpoints reflect the most stable releases as of Feb 2025. \\

\textbf{Model Roles in WiseMind }
& RA: GPT-4o, Mistral-Large, or UltraMedical  
(optimized for structured reasoning).  
\newline
EA: Claude-3.5 or EmoLLM  
(optimized for empathetic dialogue).  
\newline
Same-base experiments assign a single model to both agents; mixed-base assigns 
role-aligned LLMs based on performance specialization. \\

\textbf{Inference Settings} 
& RA: Temperature = 0.2 (deterministic reasoning).  
\newline
EA: Temperature = 0.5--0.6 (natural conversational variability).  
\newline
VSP Simulation: Temperature = 0.2 to ensure fidelity to ground-truth symptom profiles.  
\newline
All models use top-p = 1.0 unless noted. \\

\textbf{Hardware Infrastructure }
& 2$\times$ NVIDIA RTX A6000 GPUs;  
AMD Ryzen Threadripper PRO 5975WX CPU;  
256GB ECC RAM;  
6TB NVMe SSD Storage.  
Open-source models were deployed via \texttt{vLLM} on this cluster. \\

\textbf{Software Environment }
& Python 3.10; \texttt{langchain}, \texttt{langchain\_openai}, 
\texttt{langchain\_anthropic}, \texttt{langchain\_community},
\texttt{pandas}, \texttt{numpy}, \texttt{matplotlib}, \texttt{seaborn}.  
SKG retrieval and orchestration implemented using our custom multi-agent controller. \\

\textbf{Runtime and Cost }
& Average diagnostic interview: \textasciitilde{}30 seconds/turn for RA; 
\textasciitilde{}20 seconds/turn for EA (API); 
\newline
Total experiments required \textasciitilde{}20 GPU-hours and API usage cost of \$200. 
Examples provided in Supplementary Section~1.3) \\

\textbf{Reproducibility Controls }
& Deterministic decoding for reasoning steps;  
agent memory logging;  
fixed seeds for open-source runs;  
consistent prompt templates;  
risk detection and escalation disabled during benchmarking 
(see Supplementary Section~1.5). \\
\bottomrule
\end{tabular}
\end{table*}

\begin{table*}[!h]
\centering
\caption{Prompt Index for the WiseMind Framework.  
This table summarizes all prompting components used in baseline comparisons, 
multi-agent reasoning, and VSP generation. Full prompt templates are available 
in Supplementary Section 3.}
\label{tab:prompt-index}
\small
\setlength{\tabcolsep}{10pt}
\begin{tabular}{p{3cm} p{5.5cm} p{1.8cm}}
\toprule
\textbf{Prompt Type} & \textbf{Purpose / Description} & \textbf{Supplementary Location} \\
\midrule

\textbf{KFP}  
& Direct prompting without external knowledge. Used as a minimal baseline for single-agent LLM performance.  
& Supplementary Section 3.1\\

\textbf{TKEP-ICL}  
& Baseline prompting with exemplar diagnostic cases appended as demonstrations. Evaluates reasoning from textual examples.  
& Supplementary Section 3.2\\

\textbf{TKEP-RAG}  
& Baseline prompting with retrieved DSM-style textual knowledge. Tests benefit of unstructured external knowledge.  
& Supplementary Section 3.2\\

\textbf{SKEP}  
& Single-agent prompting over the SKG.  
Represents the strongest non–multi-agent baseline.  
& Supplementary Section 3.3\\

\textbf{WiseMind RA Prompt}  
& Core prompt for the Reasonable Mind Agent, responsible for 
DSM-guided reasoning, SKG traversal, and decision selection 
(\emph{Met Criteria, Not Met Criteria, More Details, Contradiction}).  
& Supplementary Section 3.3\\

\textbf{WiseMind EA Prompt}  
& Core prompt for the Emotional Mind Agent, responsible for 
empathetic interaction, rapport-building, clarifying questions, and 
patient-friendly reformulation of RA queries.  
& Supplementary Section 3.3 \\

\textbf{VSP Prompt}  
& Prompt template controlling the behavior of simulated patients, ensuring 
consistent DSM-grounded symptom portrayal during diagnostic dialogues.  
& Supplementary Section 5.1\\

\textbf{VSP Script Generation Prompt}  
& Used to generate autobiographical narratives and symptom scripts for each VSP, 
prior to dialogue-level evaluation.  
& Supplementary Section 5.1\\

\bottomrule
\end{tabular}
\end{table*}

The WiseMind system uses a generation temperature of 0.6 for doctor-agent responses. This value balances determinism with conversational naturalness—a critical requirement for psychiatric interviewing. Very low temperatures (0.0–0.3), while more deterministic, often yield repetitive or mechanical phrasing that impedes rapport~\citep{holtzman2019curious}, whereas temperatures above 0.8 may reduce coherence. Prior studies demonstrate that temperature has a limited influence on reasoning accuracy but does affect linguistic variation, suggesting that mid-range settings are appropriate for empathetic dialogue \cite{PeeperkornK0J24}. Practitioner guidelines and empirical studies also recommend temperatures between 0.5 and 0.8 for natural conversation \cite{renze-2024-effect,korbak2023pretraining}.

Diagnostic determinism is maintained through WiseMind’s structured reasoning workflow and rule-based decision templates rather than temperature alone. For simulated patients, we use a temperature of 0.2 to ensure that symptom descriptions remain consistent with ground-truth clinical profiles. Higher temperatures introduced inconsistent or hallucinated symptoms during preliminary testing. A low temperature preserves fidelity while allowing mild linguistic variability for realism.

WiseMind relies on a set of structured prompt templates for baseline methods, 
multi-agent reasoning, and VSP simulation. Table~\ref{tab:prompt-index} provides 
an index of all prompt types used in this study, while the sample prompt text 
and orchestration examples are provided in the Supplementary Section 3.

\subsection*{Dataset and Ethics}

The evaluation of WiseMind relies on interaction-centered data rather than static, label-based datasets, reflecting the multi-turn and context-dependent nature of psychiatric diagnostic interviews. As summarized in Table~\ref{tab:dataset-overview}, we developed a three-part evaluation corpus designed for rigorous and reproducible assessment: (i) the Virtual Standard Patient (VSP) Corpus for controlled benchmarking, (ii) the Simulated Interaction Corpus between VSP and WiseMind for large-scale quantitative evaluation across diverse conversational conditions, and (iii) Real Human Interaction Corpus between WiseMind and users for ecological validation and post-hoc auditing. All procedures involving human participants who enrolled at the University of Alberta were reviewed and approved by the University of Alberta Research Ethics Board (Pro00152569). Informed consent was obtained from all subjects involved in the study. All participant data was anonymized prior to analysis to ensure privacy. Below, we describe the process of constructing the VSP corpus in detail.

The VSP corpus provides controlled, interpretable, and clinically grounded diagnostic scenarios for evaluating WiseMind. VSPs parallel standardized patients in medical education by offering reproducible, ethically safe diagnostic pathways paired with coherent autobiographical narratives. Two complementary classes of VSPs were constructed.

The first class includes \textit{Human-grounded VSPs.}  
Using insights from de-identified clinical transcripts, we developed symptom-consistent autobiographical stories reflecting naturalistic patient language, emotional tone, and temporal structure. These transcripts were used solely as qualitative stylistic references: no text segments, demographic attributes, or annotations were copied, paraphrased, or used in model inputs. All identifiers were removed by the data provider. These VSPs ensure that the generated narratives align with real-world linguistic patterns without violating data-protection constraints.

The second class encompasses \textit{DSM-driven VSPs (primary dataset).}  
The primary VSP corpus was generated directly from DSM–5–TR differential-diagnosis decision pathways spanning depressive, bipolar, and anxiety disorders. For each terminal diagnosis, we extracted its full root-to-leaf diagnostic path and converted it into a structured vector of binary indicators specifying which DSM criteria were \textit{met} versus \textit{not met}. Each diagnostic node was paired with a semantic description corresponding to a clinically interpretable symptom domain (e.g., mood disturbance, sleep changes, energy level, psychomotor activity, cognition). These node descriptions serve as anchors ensuring interpretability and semantic fidelity across VSP stories. In particular, DSM-driven profiles are converted into naturalistic patient narratives using a five-step LLM pipeline:

For each terminal diagnosis, we extracted the DSM differential-diagnosis structure and formalized it into a sequence of diagnostic nodes. Each node includes (i) a binary \texttt{Met\_Criteria} value and (ii) a textual description summarizing the corresponding symptom construct. This path constitutes the complete diagnostic structure of the VSP.

Each diagnostic node $n$ was associated with a concise semantic description that identifies the clinical domain being assessed (e.g., persistent low mood, increased goal-directed activity, excessive worry, panic symptoms). These descriptions ensure that the generated narratives remain clinically meaningful and interpretable.

Before generating any symptom narratives, the LLM was instructed to internally construct a unified patient persona—including demographic background, occupation, relationship structure, cultural context, chronic stressors, and a coherent timeline of symptom evolution. This persona is not directly output; instead, it constrains all subsequent node-level narratives to ensure consistency across symptoms and prevent story-level contradictions.

Using the full diagnostic path, its \texttt{Met\_Criteria} vector, and the node descriptions, the LLM generated 2--5 first-person sentences for each diagnostic node. If \texttt{Met\_Criteria = true}, the narrative must naturally endorse and exemplify the symptom; If \texttt{Met\_Criteria = false}, the narrative must gently indicate absence of the symptom while maintaining tone and persona consistency.  
A separate LLM pass performed global consistency checking to ensure temporal, emotional, and causal coherence across all node-level stories.
  
The JSON output from the generation pipeline was parsed into a tabular structure in which each row corresponds to one diagnostic node and its corresponding symptom story. This structured profile functions as the VSP’s “internal truth.” During diagnostic simulations, the VSP responds strictly according to this profile.
 
During simulated interviews, each VSP follows a constrained role-playing protocol: responses are naturalistic but concise (1--3 sentences), grounded strictly in the available symptom narratives, and limited to answering only what the clinician explicitly asks. VSPs never volunteer information prematurely and never reference DSM concepts, internal criteria, or diagnostic labels. More details about examples and prompts can be found in Supplementary Section 5.1 and Supplementary Table 1.

\section*{Data Availability}
The datasets generated and/or analyzed during the current study are not publicly available due to requirements stated in the Research Ethics Board (REB) agreement, but are available from the corresponding author upon reasonable request. The simulated interview sessions are available at \url{https://github.com/YWU99u/WiseMind-DDx-Psyc}.

\section*{Code Availability}
The underlying code for this study is not publicly available for proprietary reasons. However, a secure test link that exposes the model behavior without revealing proprietary components can be provided by the corresponding author upon reasonable request for academic, non-commercial evaluation. 

\section*{Acknowledgment}
This study was funded by the by the National Natural Science Foundation of China (Grant No. 62576101) and the Alberta Government Department of Labour through the Supporting Psychological Health in First Responders Grant (Grant No. 22SPHIFR25-2). Y.W. gratefully acknowledges the generous financial support provided by the China Scholarship Council under the grant IDs 202308180002 (YW). The funders played no role in study design, data collection, analysis and interpretation of data, or the writing of this manuscript.

\section*{Author Contribution}
Y.W. served as the lead contributor, responsible for initiating the research ideation, leading the study design and experimental execution, all programming aspects, drafting the initial manuscript, and manuscript editing.

\noindent G.W. participated substantially in the research ideation, contributed to the study design, experiment design, programming aspects and assisted with manuscript construction.

\noindent J.L. conceived and led the research ideation and research design, proposed system architecture and evaluation strategy, and provided substantially support for manuscript drafting and editing throughout the submission and revision process.

\noindent Y.Z. contributed substantially to the study design, provided key medical insights and expertise, REB Approval, and participated in manuscript editing.

\noindent J.C. served as the Principal Investigator, coordinating the overall study, providing essential research funding, and contributing significantly to manuscript editing.

\noindent S.Z., L.M. and T.Y. participated in study discussion and contributed to the evaluation process.

\noindent M.Z. and I.P. participated in the medical insights discussion and contributed to the final evaluation of the system.

\section*{Competing Interests}
Authors Y.W., S.Z., Y.Z., and J.C. are employees of and hold equity in Shanghai KeyLinkAI Inc., which is developing commercial applications for the technology described in this manuscript. Additionally, Authors Y.W., G.W., J.L., Y.Z., and J.C. are inventors on a pending patent application regarding this technology. The remaining authors declare no competing interests.


\bibliography{sn-bibliography}

\end{document}